\documentclass[preprint,1pt,times,twocolumn]{elsarticle}

\usepackage{stfloats} 
\usepackage{lineno}

\usepackage{amssymb}
\usepackage{amsmath}
\usepackage{url}

\usepackage{graphicx}%
\usepackage{multirow}%
\usepackage{amsmath,amssymb,amsfonts}%
\usepackage{mathrsfs}%
\usepackage{xcolor}%
\usepackage{textcomp}%
\usepackage{manyfoot}%
\usepackage{booktabs}%
\usepackage{algorithm}%
\usepackage{algorithmicx}%
\usepackage{algpseudocode}%
\usepackage{listings}%
\usepackage{hyperref}
\usepackage{url}
\usepackage{enumitem}
\usepackage{wrapfig}
\usepackage{color}
\usepackage{colortbl}
\usepackage{multirow}
\usepackage{bbm}
\usepackage{nicefrac}

\def\eqref#1{equation~\ref{#1}}

\def\1{\bm{1}}

\DeclareMathAlphabet{\mathsfit}{\encodingdefault}{\sfdefault}{m}{sl}
\SetMathAlphabet{\mathsfit}{bold}{\encodingdefault}{\sfdefault}{bx}{n}

\DeclareMathOperator*{\argmax}{arg\,max}

\DeclareMathOperator{\sign}{sign}

\usepackage{siunitx}
\newrobustcmd\B{\DeclareFontSeriesDefault[rm]{bf}{b}\bfseries}

\newcommand{\myparagraph}[1]{\smallskip\noindent \textbf{#1}}

\newcommand{\vct}[1]{\ensuremath{\boldsymbol{#1}}}

\renewcommand{\argmax}{\operatornamewithlimits{\arg\,\max}}

\renewcommand{\sign}{\text{sign}}

\def \ie {\emph{i.e.}}
\def \eg {\emph{e.g.}}

\definecolor{ourcolor}{rgb}{0.87, 0.94, 0.84}
\definecolor{GrayLight}{gray}{0.92}

\author[1,2]{Hicham Eddoubi\corref{cor1}}
\ead{hicham.eddoubi@unica.it}
\cortext[cor1]{Corresponding author}

\author[3]{Jonas Ricker}
\ead{jonas.ricker@rub.de}

\author[4,5]{Federico Cocchi}
\ead{federico.cocchi@unimore.it}

\author[1,6]{Angelo Sotgiu}
\ead{angelo.sotgiu@unica.it}

\author[1,6]{Maura Pintor}
\ead{maura.pintor@unica.it}

\author[4]{Marcella Cornia}
\ead{marcella.cornia@unimore.it}

\author[5]{Lorenzo Baraldi}
\ead{lorenzo.baraldi@unimore.it}

\author[3]{Asja Fischer}
\ead{asja.fischer@rub.de}

\author[1,6]{Battista Biggio}
\ead{battista.biggio@unica.it}

\affiliation[1]{organization={University of Cagliari}, country={Italy}}
\affiliation[2]{organization={Sapienza University of Rome}, country={Italy}}
\affiliation[3]{organization={Ruhr University Bochum}, country={Germany}}
\affiliation[4]{organization={University of Modena and Reggio Emilia}, country={Italy}}
\affiliation[5]{organization={University of Pisa}, country={Italy}}
\affiliation[6]{organization={CINI}, country={Italy}}

\makeatletter
\AtBeginDocument{\def\maketitle{\elsarticleprelims\twocolumn[\finalMaketitle]\printFirstPageNotes\gdef\thefootnote{\arabic{footnote}}}}
\makeatother

\begin{document}

\begin{frontmatter}

\title{RAID: A Benchmark Dataset for Testing the Adversarial Robustness of AI-Generated Image Detectors}

\begin{abstract}

AI-generated images (AIGIs) have reached a level of quality at which humans are unable to reliably distinguish them from real images. To counter the inherent risks of fraud and disinformation, their detection is a pressing challenge and an active research topic.
While many of the presented methods claim high detection accuracy, they are often evaluated without considering their \textit{adversarial robustness}.
In real-world scenarios, malicious actors can modify AIGIs, e.g., through imperceptible perturbations, to evade detection systems. In this work, we provide a dataset to assess the robustness of AI-generated image detectors. We present \textbf{RAID} (\textbf{\underline{R}}obust evaluation of \textbf{\underline{AI}}-generated image \textbf{\underline{D}}etectors), a benchmark dataset of 216k diverse and transferable adversarial images.
The proposed dataset is created starting from images generated by four different text-to-image models (to promote diversity) and optimizing attacks against an ensemble of seven state-of-the-art detectors (to promote transferability).
Extensive experiments show that our methodology generates adversarial images that transfer to unseen detectors, degrading the performance of every detector outside the attack ensemble that we evaluate, and provides a practical, worst-case robustness screen: failure on RAID is a strong indicator that a detector is not adversarially robust.

\end{abstract}


\end{frontmatter}


\section{Introduction}

In recent years, generative artificial intelligence has evolved from a mere research topic to a vast collection of commonly available tools.
While improving tasks such as image super-resolution~\cite{Wang_2024_CVPR} and dataset augmentation~\cite{10943975}, this rapid development poses considerable risks of misuse leading to misinformation spread, privacy violation, and identity theft~\citep{rickerAIgeneratedFacesReal2024}. This underscores the urgent need for detection methods that generalize and keep up with ever-evolving image generation technology.
To mitigate the harmful consequences of AI-generated images (AIGIs), a variety of detection approaches have been proposed in the literature~\citep{wang2020cnn,guoEyesTellAll2022,ojha2023towards,chen2024drct,CoDE2024,wangDIREDiffusiongeneratedImage2023}.
Based on the reported near-perfect accuracy, the problem of detecting AIGIs appears to have already been solved.
However, evaluations are typically conducted without considering real-world risks. One major factor is the effect of common processing operations, such as resizing or compression, which have already been shown to have drastic effects on the AIGI detectors' performance~\citep{gragnanielloAreGANGenerated2021,xuProFakeDetectingDeepfakes2024,cocchi2023unveiling}.\\

\begin{figure*}[t]
  \centering
  \includegraphics[width=\textwidth]{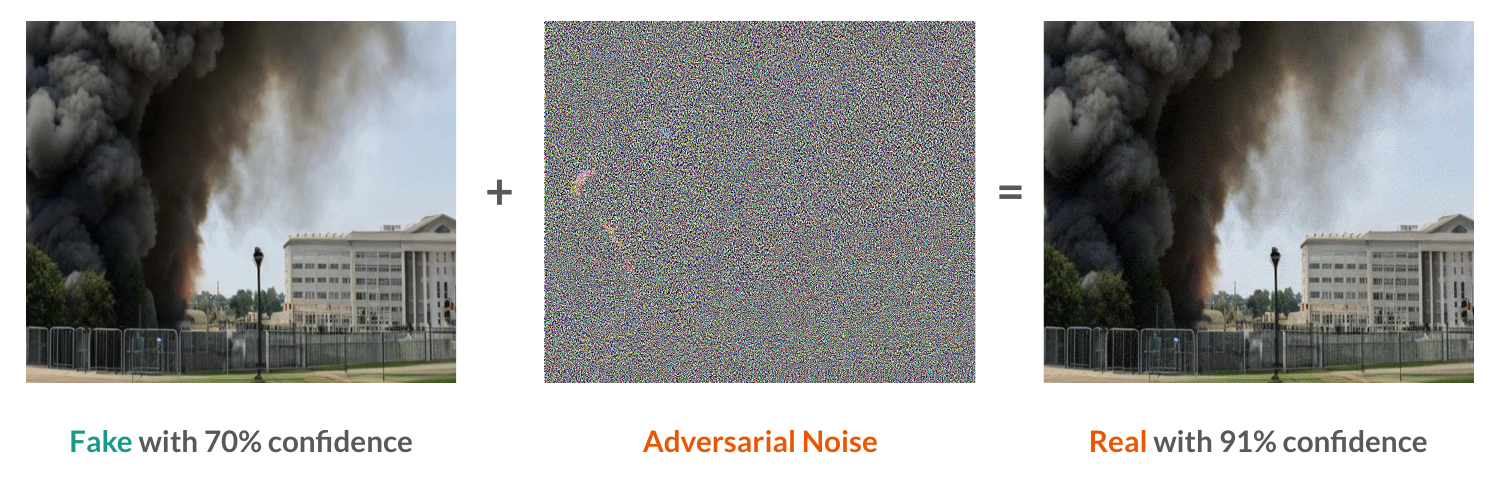}
  \vspace{-0.4cm}
  \caption{Adversarial attack on the AIGI detector provided by \textit{sightengine}. Successful evasion is shown by adding adversarial noise computed on the detector by~\citet{corvi2023detection}. We perform an extended evaluation on commercial detectors in \ref{app:commercial}.}
  \label{fig:evasion_example}
\end{figure*}

Another important factor that the majority of existing work has neglected is the \textit{adversarial robustness} of detectors.  In fact, attackers can introduce carefully crafted, often imperceptible modifications to AIGIs and systematically bypass detection without degrading image quality.
Take, for instance, \autoref{fig:evasion_example}, an AIGI that was spread across social media outlets in 2023~\citep{2023pentagon} that, despite being quickly debunked as fake, still had an impact on the stock market. We can detect such an image as AI-generated with an off-the-shelf detector, but when we modify the image with adversarial perturbations, it can evade detection by that detector and others, with increased effectiveness against those with a similar architecture~\citep{mavali2024fake}. 
The adversarial robustness of AIGI detectors is often not investigated, partly because generating adversarial images requires significant effort. Due to the many attack algorithms, hyperparameters, and the required technical knowledge, conducting a comprehensive robustness analysis is not straightforward.

Existing work~\citep{mavali2024fake} reveals the lack of robustness of AIGI detectors in white-box scenarios, where the malicious actor has access to the model's architecture and training parameters, and in black-box scenarios, where the attacker's knowledge is limited. However, we note that the attacks used for the evaluation remain restricted to techniques that increase their success and transferability.
In this work, we extend this concept to bridge this evaluation gap by providing a dataset to assess the adversarial robustness of AIGI detectors. In particular, we propose \textbf{RAID} (\textbf{\underline{R}}obust evaluation of \textbf{\underline{AI}}-generated image \textbf{\underline{D}}etectors), a large-scale benchmark dataset of \textit{diverse} and \textit{transferable} adversarial images. To promote diversity and transferability, we use images generated by four different text-to-image models and optimize attacks against an ensemble of state-of-the-art AIGI detectors with different architectures. As we demonstrate experimentally, testing detection performance on RAID provides a fast and necessary robustness screen: a detector that fails on RAID is very unlikely to be robust to stronger, detector-specific attacks. Our benchmark across seven recently proposed detectors shows that the current landscape of AIGI detection is neither expansive nor reliable enough for widespread real-world adoption without proper adversarial robustness.

\myparagraph{Contributions.} In summary, we make the following contributions:
\begin{itemize}[noitemsep,nolistsep]
    \item We create RAID, the first benchmark dataset of transferable adversarial images to test the adversarial robustness of state-of-the-art AIGI detectors, released in three splits generated with different ensemble attacks (PGD, APGD, and CWA), enabling reproducible and comparable assessment.
    \item We conduct a large-scale study across diverse AIGI detectors, showing that black-box adversarial images evade detection in both open-source and commercial closed-source detectors, and that no current detector is reliably robust to transferable adversarial attacks.
    \item We show that the RAID's adversarial images provide a fast robustness screen without white-box access to the target model, making it a practical drop-in robustness test for new detectors: failure on RAID implies a lack of adversarial robustness.
\end{itemize}
\section{Adversarial Problem Definition}
\label{crafting_adv_perturbations}
\myparagraph{AIGI Detection.}
We consider the problem of detecting AIGIs as a binary classification task. 
Let $\vct x \in [0, 1]^d$ denote an input image and $y \in \{0, 1\}$ be the corresponding ground-truth label, \ie, 1 for AIGIs and 0 for real images.
An AIGI detector can be modeled as a function $f_{\vct \theta}: [0, 1]^d \to [0, 1]$ parametrized by $\vct \theta$, whose output represent the probability that $\vct x$ is AI-generated. The predicted label $\hat y$ can then be obtained by applying a fixed threshold $\tau$, such that a sample is considered an AIGI if $f_{\vct \theta}(\vct x) \geq \tau$, and a real image otherwise.

This formulation is shared by most state-of-the-art AIGI detectors, independently of their underlying architecture.

\myparagraph{Crafting Adversarial Examples.}
Given an image with ground-truth label $y \in \{0, 1\}$, the goal of an adversarial attack is to produce a perturbation that makes the detector misclassify it, while ensuring that the perturbation does not significantly corrupt the image. 
We therefore cast evasion as an \emph{untargeted} attack: the attacker seeks a perturbation $\vct \delta \in \mathbb R^d$ maximizing a loss representing the attacker's objective, solving the following optimization problem~\citep{biggio13-ecml, szegedy14-iclr}:
\begin{equation}
    \argmax_{\vct \delta : \| \vct \delta \|_{\infty} \leq \epsilon} \ \mathcal{L} \big(\vct x + \vct \delta,\, y, \vct \theta\big),
    \label{eq:evasion_attack}
\end{equation}
where $\epsilon$ bounds the perturbation size in the $\ell_\infty$ norm, and $\mathcal{L}$, in our case, is the cross-entropy $\mathcal{L}_{CE}$ with respect to the true class $y$.
The optimization problem in \autoref{eq:evasion_attack} is commonly solved using gradient-based techniques, which assume full access to the detector’s parameters and gradients. 
This corresponds to a \emph{white-box} attack setting. 
In practice, however, such access is rarely available, as deployed detection systems are typically proprietary or accessible only through black-box interfaces.
Despite this limitation, adversarial perturbations crafted against a specific model can still induce misclassification on other, unseen detectors. 
This phenomenon, known as \emph{transferability}, enables \emph{black-box} attacks in which the adversary optimizes perturbations on surrogate models and deploys them against a target detector. 
While transferability is generally weaker across heterogeneous architectures, it can be substantially improved by jointly attacking multiple models~\citep{liu2017delving,dong2018boosting}.

\myparagraph{Promoting Transferability.} To increase transferability, we extend \autoref{eq:evasion_attack} to an ensemble of $M$ detectors with parameters $\{\vct \theta_m\}_{m=1}^M$ with an ensemble loss computed as the average of the individual detectors' losses:
\begin{equation}
    \argmax_{\vct \delta : \| \vct \delta \|_{\infty} \leq \epsilon} \quad
     \frac{1}{M} \sum_{m=1}^M \mathcal{L}_{CE}\!\left(\vct x + \vct \delta,\, y, \vct \theta_m\right) \, .
    \label{eq:ensemble_attack}
\end{equation}

We solve \autoref{eq:ensemble_attack} with (untargeted) Projected Gradient Descent (PGD)~\citep{madry2018towards}. Algorithm~\ref{alg:ensemble_pgd} describes the complete procedure in detail.
By simultaneously optimizing against a diverse set of detectors, ensemble-based attacks encourage perturbations that exploit weaknesses shared across models, rather than detector-specific artifacts. 
As a result, such perturbations exhibit significantly higher transferability to unseen detectors. 
From an evaluation perspective, \textit{if a detector fails under strong transferable adversarial attacks, it is in practice unlikely to be robust against more powerful attacks}, including fully white-box ones. 
This observation motivates the construction of transferable adversarial benchmarks for robustness assessment.

\myparagraph{Threat Model.}
To avoid ambiguity between the experiments we run in the paper and the released dataset, we make the access assumptions explicit. We distinguish two settings: \emph{white-box}, where the attacker has full access to the target detector's architecture, parameters, and gradients; and \emph{black-box} (\emph{transfer}), where the target detector is excluded from the surrogate ensemble and is unknown to the attacker. \autoref{tab:threat_model} maps every experiment in the paper to its setting, the surrogate ensemble used, and whether the corresponding images are part of the public RAID release. The first six listed experiments are target-specific diagnostics and are not part of the release, while the released RAID dataset is generated with the full seven-detector ensemble and is evaluated only on detectors outside that ensemble.

\begin{table}[t]
\centering
\caption{Threat model for each experiment: access setting, surrogate ensemble, and whether the images belong to the public RAID release. $N$ is the seven-detector pool of \autoref{detection_ensemble}.}
\label{tab:threat_model}
\setlength{\tabcolsep}{.3em}
\renewcommand{\arraystretch}{1.15}
\small
\begin{tabular}{@{}>{\raggedright\arraybackslash}p{.37\columnwidth}l>{\raggedright\arraybackslash}p{.24\columnwidth}c@{}}
\toprule
\textbf{Experiment} & \textbf{Setting} & \textbf{Surrogate ensemble} & \textbf{In release?} \\
\midrule
PGD Hyperparameters Ablation \newline(\autoref{fig:asr_vs_config}) & black-box & $N \setminus \{\text{target}\}$ & no \\
Single-detector attack \newline(Tabs.~\ref{tab:whitebox_eval_16}, \ref{tab:whitebox_eval_32}, \ref{tab:whitebox_eval_8}, diag.) & white-box & target only & no \\
Cross-model transfer \newline(Tabs.~\ref{tab:whitebox_eval_16}, \ref{tab:whitebox_eval_32}, \ref{tab:whitebox_eval_8}, off-diag.) & black-box & one source detector & no \\
Leave-one-out ensemble \newline(Tabs.~\ref{tab:whitebox_eval_16}, \ref{tab:whitebox_eval_32}, \ref{tab:whitebox_eval_8}, last row) & black-box & $N \setminus \{\text{target}\}$ & no \\
Additional generators \newline(\autoref{tab:external_generators}) & black-box & $N \setminus \{\text{target}\}$ & no \\
Other ensemble attacks \newline(\autoref{tab:other_attacks}) & black-box & $N \setminus \{\text{target}\}$ & no \\
RAID -- extra detectors \newline(Tabs.~\ref{tab:raid_eval}--\ref{tab:raid_eval_extra}) & black-box & full $N$ & yes \\
Commercial detectors \newline(\ref{app:commercial}) & black-box & full $N$ & yes \\
\bottomrule
\end{tabular}
\end{table}

\section{The RAID Benchmark}  \label{RAID}
\begin{figure*}
    \centering
    \includegraphics[width=\linewidth]{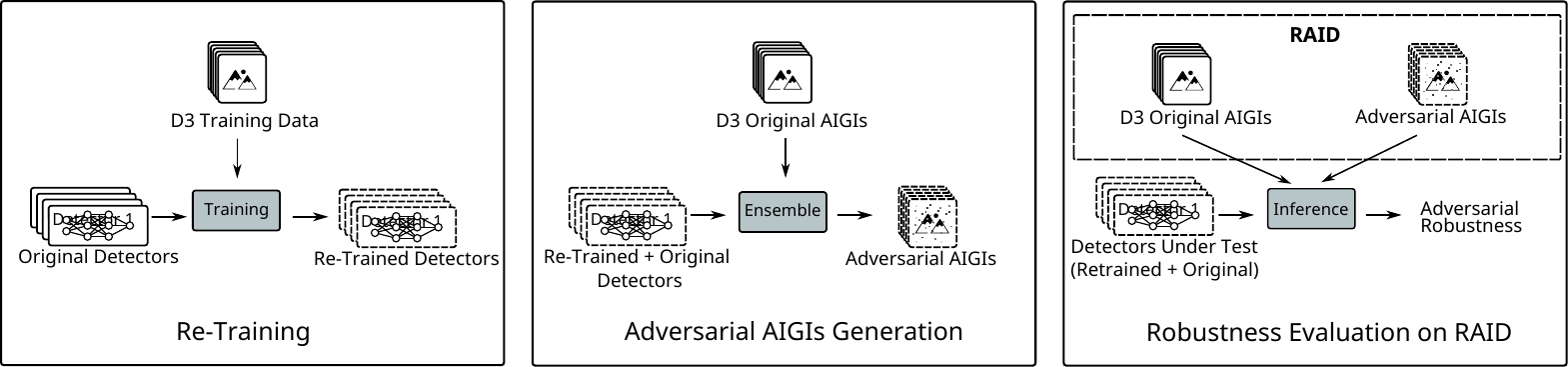}
    \vspace{-0.4cm}
    \caption{A diagram of the three experimental pipelines used in the paper for generating the RAID dataset. Left: We keep the top three-performing detectors intact while retraining the rest. Middle: We generate adversarial images optimizing on an ensemble of detectors. Right: We evaluate the baseline detectors on the RAID images.}
    \label{fig:diagram}
\end{figure*}

This section describes how we constructed our dataset of transferable adversarial images. 
First, we detail the design choices that led to the creation of RAID, namely the choice of the initial dataset and the pool of AIGI detectors involved in the study. 
We then illustrate our three-step pipeline for creating and using the dataset, shown in \autoref{fig:diagram}.

\subsection{Source Dataset}
\label{sec:source_dataset}
RAID is built upon the D\textsuperscript{3} dataset~\citep{CoDE2024}.
In total, D\textsuperscript{3} consists of 11.5M images.
It is constructed from 2.3M real images taken from the LAION-400M~\citep{schuhmannLAION400MOpenDataset2021} dataset.
Using the corresponding caption as prompts, AIGIs are generated from four open-source text-to-image models: 
Stable Diffusion v1.4~\citep{rombach2022high}, Stable Diffusion v2.1~\citep{rombach2022high}, Stable Diffusion XL~\citep{podell2023sdxl}, and DeepFloyd IF~\citep{DeepFloydIF}. 
For details on data selection and prompt engineering, we refer to the original publication~\citep{CoDE2024}. 
It should be noted that each AIGI is post-processed such that the image format and compression strength match that of the corresponding image from the distribution of real images.
This not only reduces the risk of unwanted biases between real and generated images but also makes the dataset significantly more challenging for the detection task.
We focus on diffusion-based text-to-image models, as opposed to GAN-based models, because of their higher image quality and wider use in practice~\cite{dhariwalDiffusionModelsBeat2021,bieRenAIssanceSurveyAI2025,linEvaluatingTexttoVisualGeneration2024}.

\subsection{Detection Ensemble}
\label{detection_ensemble}
The key requirement for our benchmark is that adversarial perturbations are \textit{transferable} (\ie, they are effective against unseen detectors). To achieve that, we use an ensemble of a diverse set of seven different detectors:

\myparagraph{\citet{wang2020cnn}.} It employs a ResNet-50~\citep{heDeepResidualLearning2016} trained on real and generated images from ProGAN~\citep{karrasProgressiveGrowingGANs2018}. During training, extensive data augmentation is used to account for different image processing and increase generalization.

\myparagraph{\citet{corvi2023detection}.} This architecture is adapted from \citep{gragnanielloAreGANGenerated2021} and is a modification of the detector by \citet{wang2020cnn}, that avoids downsampling in the first layer to preserve high-frequency features and applies stronger augmentation.

\myparagraph{\citet{ojha2023towards}.} It uses a pre-trained vision foundation model (CLIP~\citep{radfordLearningTransferableVisual2021}) as a feature extractor to avoid overfitting on a particular class of generated images and, thus, improve generalization. A linear classifier is added and trained on top of the 768-dimensional feature vector.

\myparagraph{\citet{koutlis2024leveraging}.} It extends~\citet{ojha2023towards} by also using intermediate encoder blocks from CLIP. The resulting features are weighted using a learnable projection network, followed by a classification head.

\myparagraph{\citet{cavia2024real}.} Instead of classifying the entire image at once, this detector operates on $9\times9$ patches. The architecture is based on ResNet-50~\citep{heDeepResidualLearning2016} but uses $1\times1$ convolutions to limit the receptive field.

\myparagraph{\citet{chen2024drct}.} It is a training paradigm that can increase the generalizability of AIGI detectors by leveraging diffusion models to reconstruct semantically similar images containing detectable artifacts. The authors provide pre-trained detectors based on two backbones, ConvNeXt~\citep{Liu_2022_CVPR} and CLIP~\citep{radfordLearningTransferableVisual2021}.

\subsection{Creating RAID} \label{subsec:creating_raid}

\autoref{fig:diagram} illustrates the overall workflow adopted in this paper for constructing and using RAID. 
The pipeline consists of three main stages: (i) re-training a subset of detectors to ensure reliable clean performance, (ii) generating transferable adversarial images via ensemble-based attacks, and (iii) evaluating the adversarial robustness of detectors on the resulting RAID benchmark. 
In the following, we describe each step in detail and provide practical guidance on assessing the robustness of AIGI detectors using RAID.

\begin{table}
\caption{Evaluation of each model on a subset of the clean D\textsuperscript{3} test set. † refers to detectors trained on D\textsuperscript{3}. F1-score computed with the AI-generated class as positive.} 
\label{tab:clean_eval}
\centering
\setlength{\tabcolsep}{.6em}
\small
\begin{tabular}{llccc}
\toprule
 && \textbf{F1} & \textbf{Acc} & \textbf{AUROC} \\
\midrule
    \cite{corvi2023detection} & (Co23) & 0.75 & 0.81 & 0.91 \\
    \cite{ojha2023towards} & (O23) & 0.33 & 0.29 & 0.68 \\
    \cite{wang2020cnn} & (W20) & 0.21 & 0.02 & 0.46 \\
    \cite{cavia2024real} & (Ca24) & 0.20 & 0.07 & 0.45 \\
\midrule
    \cite{wang2020cnn}\textsuperscript{†} &(W20\textsuperscript{†}) & 0.99 & 0.99 & 0.99 \\
    \cite{corvi2023detection}\textsuperscript{†} &(Co23\textsuperscript{†}) & 0.98 & 0.99 & 0.99 \\
    \cite{ojha2023towards}\textsuperscript{†} &(O23\textsuperscript{†}) & 0.92 & 0.95 & 0.95 \\
    \cite{chen2024drct} &(ConvNext) (Ch24CN) & 0.86 & 0.91 & 0.94 \\
    \cite{koutlis2024leveraging} &(K24) & 0.74 & 0.81 & 0.88 \\
    \cite{cavia2024real}\textsuperscript{†} &(Ca24\textsuperscript{†}) & 0.55 & 0.63 & 0.84 \\
    \cite{chen2024drct} &(CLIP) (Ch24C) & 0.81 & 0.89 & 0.73 \\
\bottomrule
\end{tabular}
\end{table}

\myparagraph{Re-training.} The first step of the pipeline (\autoref{fig:diagram}, \textit{left}) aims to ensure that the detectors used to generate adversarial perturbations exhibit adequate performance on clean images. 
As shown in \autoref{tab:clean_eval}, the published versions of most detectors do not perform well on the original images in our dataset. We hypothesize that this may be attributed to the applied post-processing and the specific test images, which differ greatly from the training data of the various detectors.
Since creating adversarial images based on detectors that are already ineffective against clean samples reduces the impact of the work, we re-train detectors on the training subset described in \autoref{sec:source_dataset}, following the original authors' training instructions. We note that the subset of re-trained detectors was chosen to maintain a practical balance between the diversity of the training data and performance on clean samples, thereby avoiding dataset bias.
As shown in \autoref{experimental_results}, retraining ensures sufficient performance and, subsequently, the generation of effective adversarial images.

\myparagraph{Adversarial Images Generation.} 
In the second step (\autoref{fig:diagram}, \textit{middle}), we generate adversarial images by optimizing perturbations against an ensemble of detectors.
To create the initial dataset, we use the same procedure as in~\citep{CoDE2024} to generate AIGIs from 4,800 new real images, obtaining $4{,}800 \times 4 = 19{,}200$ AIGIs (four generators per real image) alongside the $4{,}800$ real images, for $24{,}000$ unique images in total.
Following the untargeted formulation of \autoref{crafting_adv_perturbations}, we perturb images of \emph{both} classes: each of the $24{,}000$ images is attacked at each budget $\epsilon$ (real images are themselves perturbed, and no image is duplicated across splits).
The released benchmark comprises three attack splits, generated with PGD, APGD~\citep{pmlr-v119-croce20b}, and CWA~\citep{chen2023rethinking} using the same ensemble, budgets, and hyperparameters (\ref{supp:other_attacks}), each covering the three budgets $\epsilon \in \{\frac{8}{255}, \frac{16}{255}, \frac{32}{255}\}$. Each split contains $24{,}000 \times 3 = 72{,}000$ adversarial images -- of which $19{,}200 \times 3 = 57{,}600$ are adversarial AI-generated images and $4{,}800 \times 3 = 14{,}400$ are adversarial real images -- for $216{,}000$ adversarial images in total, released together with the $24{,}000$ original images. The PGD split is the main split of RAID, on which all experiments in this paper are based. The dataset documentation reports the exact class distribution of every split.

\myparagraph{Robustness Evaluation on RAID.}
The final step of the pipeline (\autoref{fig:diagram}, \textit{right}) consists of evaluating AIGI detectors on the RAID benchmark. 
This evaluation additionally serves as a practical tutorial on how RAID should be used to evaluate the robustness of new AIGI detectors.
Given a detector under test, robustness is assessed by measuring its detection performance on adversarial images. 
Since adversarial perturbations are generated using an ensemble that excludes the detector under evaluation, this setting naturally corresponds to a transferable black-box attack.
A significant performance drop on RAID indicates high vulnerability.
\textit{Importantly, failure under this evaluation suggests that the detector is unlikely to remain reliable in real-world adversarial settings, where attackers may have access to even stronger, detector-specific attacks.}
We emphasize that RAID provides a \emph{one-directional} (necessary, not sufficient) screen: failure on RAID implies a lack of adversarial robustness, whereas maintaining high performance on RAID is a favorable but not conclusive signal, since a detector could still be vulnerable to a detector-specific white-box attack or to a different transfer distribution. RAID should therefore be reported as a strong lower bound on adversarial vulnerability rather than as a complete characterization of robustness.
\section{Experimental Analysis} \label{experimental_analysis}
In this section, we show how RAID can be used to estimate the adversarial robustness of AIGI detectors. To this end, we initially test the performance on unperturbed images and conduct classical white-box attacks on each detector, demonstrating their susceptibility to adversarial images. We subsequently analyze the transferability of white-box attacks and compare them to our ensemble attack, showing the effectiveness of RAID as an easy-to-use tool for assessing adversarial robustness.

\subsection{Experimental Setup} \label{experimental_setup}
\myparagraph{Detectors.} In addition to the seven detectors described in \autoref{detection_ensemble}, we build four additional detectors by training a linear classifier on top of four pre-trained visual backbones, using the D\textsuperscript{3} train set.
In particular, we use two different versions of DINO~\citep{caron2021emerging,darcet2023vision} to highlight the behavior of self-supervised models. At the same time, to explore the impact of model size, we also consider a ViT-Tiny.
Additionally, we adapt CoDE~\citep{CoDE2024} for this analysis. Specifically, we use the feature extractor trained with a contrastive loss tailored for AIGI detection and train a linear classification layer on top. For DINO and ViT-Tiny, we use the transformation pipeline introduced by~\cite{ojha2023towards} for training and evaluation, whereas for CoDE, we use the transformation strategy described in its original implementation. This setup enables consistent comparisons across different architectures and training paradigms.

\myparagraph{Dataset.} As noted in \autoref{RAID}, we consider the images of the D\textsuperscript{3} test data set. Following the approach used for D\textsuperscript{3}, only images labeled \textit{safe} in the LAION metadata are considered real. This ensures that the generated images also adhere to this safeguard. To this end, samples depicting NSFW images have been excluded. \textit{All images are center-cropped} to $256\times256$ rather than resized, before applying the detector-specific pre-processing and evaluation: down-sampling smooths local pixel correlations and can distort the low-level artifacts that detectors rely on, whereas cropping preserves them~\citep{li2025safe,li2025aigibench}. A fixed crop size also provides consistent input dimensions for efficient batch processing, and the perturbation is optimized over the entire $256\times256$ crop, which constitutes the released image. During dataset construction, we save the adversarial images as PNGs, avoiding lossy formats (\eg, JPEG), since we only consider the worst-case adversarial scenario in which no post-processing operations, which could reduce the attack's transferability~\cite{guo2018countering}, are performed.

\myparagraph{Evaluation Protocol.} Initially, we use all the 24,000 unperturbed images to evaluate the detectors' performance in the absence of adversarial attacks. We then select a subset of 1,000 test images per class (5,000 in total) and use them to conduct a thorough evaluation of the detector's performance with and without adversarial attacks across several settings. First, we assess the adversarial robustness of detectors in a white-box setting where each detector is attacked using adversarial perturbations crafted against it. This scenario represents the worst-case scenario in which the attacker has full knowledge of the target detector's architecture and parameters. Furthermore, we investigate the cross-model transferability of adversarial images computed against each detector to the other six detectors, examining whether adversarial perturbations designed to fool one detector can also fool the others. Next, we apply the ensemble attack approach described in \autoref{crafting_adv_perturbations} in a \textit{leave-one-out manner}, \ie, excluding the assessed detector from the attacked ensemble. This ensemble approach enables us to evaluate both the transferability of ensemble attacks and the detector's performance in a black-box scenario, where access to the targeted detector architecture or parameters is unavailable. Finally, to assess transfer to genuinely unseen detectors whose architectures and training data are independent of our ensemble, we evaluate five recent state-of-the-art detectors: Effort~\citep{effort}, AIDE~\citep{yanSanityCheckAIgenerated2025b}, OmniAID~\citep{omniaid}, DDA~\citep{dda}, and DDD~\citep{yang2025d3}, directly on the pre-computed RAID dataset. 

Additionally, to demonstrate the adaptability of our method and broaden the evaluation scope, we apply our attack methodology to a set of 11 additional text-to-image models: Kandinsky 2.1, Kandinsky 2.2, Kandinsky 3~\citep{razzhigaev2023kandinsky}, aMUSEd~\citep{patil2024amused}, PixArt-$\alpha$~\citep{chen2023pixart}, SD-XL-Turbo~\citep{podell2023sdxl}, SAG~\citep{hong2023improving}, unCLIP~\citep{rameshHierarchicalTextconditionalImage2022}, MultiDiffusion~\citep{bartal2023multidiffusion}, and the latent consistency models LCM~\citep{luo2023lcm} and LCM-XL~\citep{luo2023lcmlora}.
The selected models span a diverse range of architectures to test the generalizability of our approach under different conditions.
We generate synthetic counterparts to the 4,800 real images in our benchmark, yielding a total of 52,800 AIGIs, and apply our leave-one-out ensemble attack to a subset of 200 images per generator, together with 200 real images, varying the perturbation budget.    

Finally, we construct the RAID dataset, as shown in \autoref{fig:diagram}, by running the adversarial attack on the ensemble of seven detectors using the entire D\textsuperscript{3} test set, and then evaluate its effectiveness on nine additional detectors.

\myparagraph{Attack Setting.} The attack optimization is performed with the Projected Gradient Descent (PGD) adversarial attack~\citep{madry2018towards}\footnote{The released RAID additionally includes splits generated with APGD~\citep{pmlr-v119-croce20b} and CWA~\citep{chen2023rethinking}, using the same ensemble, budgets, and hyperparameters; all experiments in this paper use the PGD split. \ref{supp:other_attacks} compares the three attacks.} (Algorithm~\ref{alg:ensemble_pgd}), a well-established iterative approach to generate adversarial examples. We employ PGD with iterations fixed at 10, as a longer optimization tends to overfit the surrogate models and reduce transferability~\cite{demontis2019why,xiong2022svre,chen2023rethinking}, zero initialization, and a budget-dependent step size, selected with the ablation of \autoref{sec:pgd_ablation}: $\alpha = 0.01$ for $\epsilon = \frac{8}{255}$, and $\alpha = 0.03$ for $\epsilon = \frac{16}{255}$ and $\epsilon = \frac{32}{255}$\footnote{The results for the perturbation $\epsilon = \frac{8}{255}$ are reported in \ref{supp:additional_results}.}.

\myparagraph{Experimental Environment.}
During the detectors' retraining phase, 4 A100 GPUs are used in a distributed data-parallel setup, requiring each up to 18 hours until convergence.
In contrast, the ensemble attack is run on a single A100 GPU and takes 8 hours. The evaluation script is lightweight and takes less than 1 hour for each detector in a non-distributed setting.

\myparagraph{Metrics.} We restrict the adversarial analysis to two threshold-free measures: the AUROC and the \emph{attack success rate} (ASR), defined as the relative AUROC drop with respect to the clean images, $\text{ASR} = 100 \cdot \frac{\text{AUROC}_{\text{clean}} - \text{AUROC}_{\text{adv}}}{\text{AUROC}_{\text{clean}}}$. Accuracy and F1-score depend on the decision threshold, which is not calibrated for our data. We therefore report them only for the clean evaluation. As untargeted attacks perturb both classes, an AUROC of $0.5$ corresponds to random ranking, and $0.0$ corresponds to inverted ranking. For commercial detectors (\ref{app:commercial}), we additionally report accuracy with threshold fixed at $0.5$.

\subsection{Ablation on PGD hyperparameters}
\label{sec:pgd_ablation}

\begin{figure}[t]
    \centering
    \includegraphics[width=\linewidth]{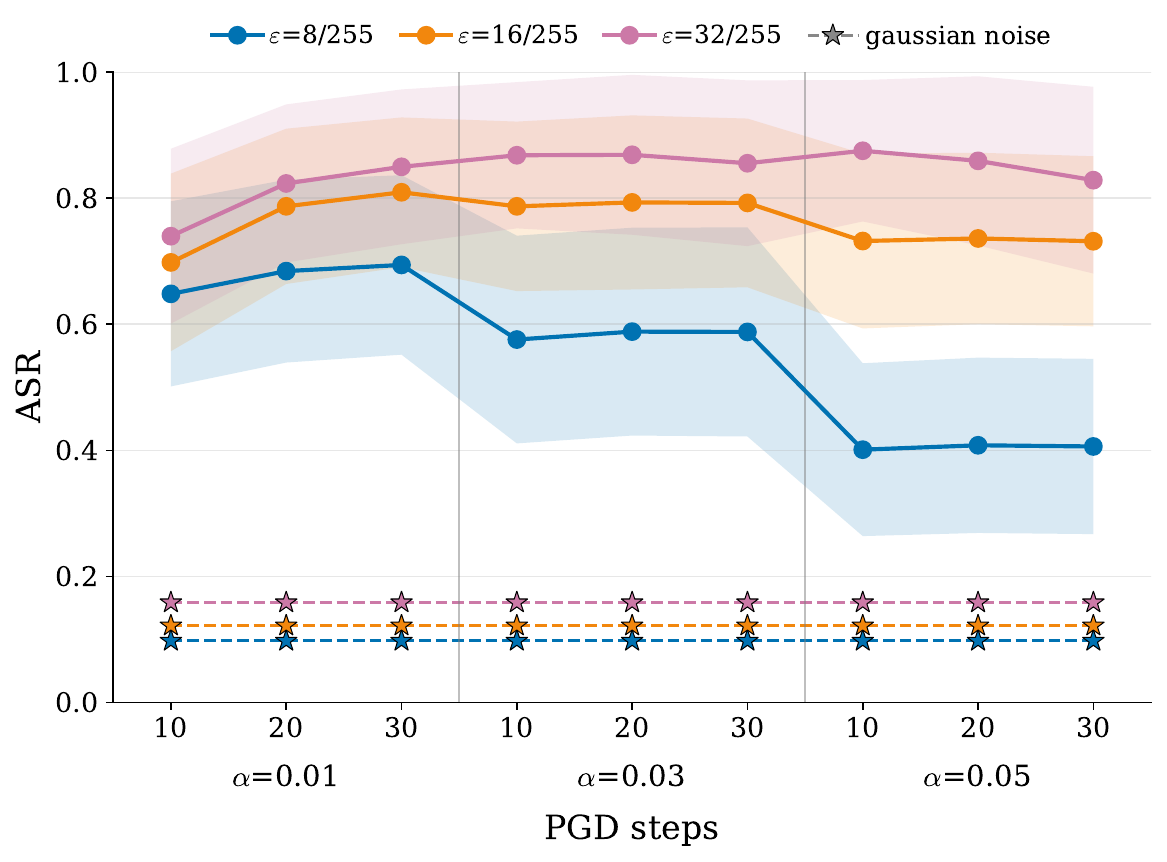}
    \vspace{-0.4em}
    \caption{ASR of the leave-one-out ensemble PGD attack for every number of iterations $K$ (ticks) and step size $\alpha$ (groups), for each budget $\epsilon$ (colors). Dashed lines: Gaussian noise clipped to the same $\ell_\infty$ budget. Mean over 3 random starts and the 7 held-out detectors, on 1,000 images; shaded areas: $\pm1$ standard deviation.}
    \label{fig:asr_vs_config}
\end{figure}

We select the attack configuration with an ablation of the leave-one-out ensemble attack on a subset of 1,000 test images (200 real images and 200 AIGIs from each of the four generators). For every budget, we vary the number of iterations $K\in\{10,20,30\}$ and the step size $\alpha\in\{0.01,0.03,0.05\}$, with three seeded random starts per configuration, and report the ASR averaged over the random starts and the seven held-out detectors in \autoref{fig:asr_vs_config}. As a benign-distortion control, we add Gaussian noise with standard deviation $\epsilon$, clipped to $[-\epsilon,\epsilon]$.
With $K=10$, the best step size depends on the budget: $\alpha=0.01$ at $\epsilon=\frac{8}{255}$ (ASR 64.8\%, against 57.6\% and 40.1\% for $\alpha=0.03$ and $0.05$, whose steps are comparable to or larger than the budget itself), and $\alpha=0.03$ at $\frac{16}{255}$ (78.7\%, against 69.8\% and 73.2\%). At $\frac{32}{255}$, $\alpha=0.03$ and $0.05$ are within one point (86.8\% and 87.5\%), and we keep $\alpha=0.03$. Increasing $K$ from 10 to 30 changes the ASR of the selected configurations by at most 4.6 points for three times the computation, so we use $K=10$ for all budgets. Gaussian noise of the same $\ell_\infty$ norm yields an ASR of only 9.8--15.9\%, against 64.8--86.8\% for the attack, confirming that the degradation stems from the adversarial optimization rather than from generic image corruption.

\subsection{Perceptibility of adversarial perturbations}
\label{sec:perceptibility}

An $\ell_\infty$ bound alone does not quantify how much a perturbation distorts an image. We therefore measure LPIPS~\citep{lpips} and SSIM~\citep{ssim} between each adversarial image and its clean counterpart, for the released RAID dataset~(\autoref{tab:raid_perceptual}). Both are image-distance measures, LPIPS being learned: a lower value supports a claim of lower measured distortion, not of human imperceptibility or of a perturbation below a visibility threshold. Accordingly, we do not describe the RAID perturbations as imperceptible: $\epsilon=\frac{8}{255}$ yields the lowest measured distortion, whereas $\frac{16}{255}$ and $\frac{32}{255}$ introduce clearly measurable structural changes and should be read as bounded, worst-case perturbations.

\begin{table}[t]
\centering
\caption{LPIPS and SSIM between the adversarial images of the released RAID dataset and their clean counterparts, per perturbation budget $\epsilon$. Mean $\pm$ standard deviation over all images.}
\label{tab:raid_perceptual}
\setlength{\tabcolsep}{.25em}
\small
\begin{tabular}{ccc|ccc}
\toprule
\multicolumn{3}{c|}{\textbf{LPIPS} $\downarrow$} & \multicolumn{3}{c}{\textbf{SSIM} $\uparrow$} \\
\cmidrule(lr){1-3} \cmidrule(lr){4-6}
$\epsilon=\frac{8}{255}$ & $\epsilon=\frac{16}{255}$ & $\epsilon=\frac{32}{255}$ & $\epsilon=\frac{8}{255}$ & $\epsilon=\frac{16}{255}$ & $\epsilon=\frac{32}{255}$ \\
\midrule
0.09\scriptsize{$\pm$0.07} & 0.24\scriptsize{$\pm$0.13} & 0.40\scriptsize{$\pm$0.16} & 0.85\scriptsize{$\pm$0.07} & 0.63\scriptsize{$\pm$0.12} & 0.47\scriptsize{$\pm$0.13} \\
\bottomrule
\end{tabular}%
\end{table}

\subsection{Experimental Results} 
\label{experimental_results}

\myparagraph{Initial Evaluation.} Prior to evaluating the adversarial robustness of the considered detectors, we first evaluate their performance on the D\textsuperscript{3} test set.
As already noted in \autoref{subsec:creating_raid}, the initial performance evaluation of the detectors with the provided weights in \autoref{tab:clean_eval} shows mixed results, particularly regarding F1-score and accuracy measures.
We thus retrain a selected subset of detectors on the D\textsuperscript{3} training set: \citet{cavia2024real}, \citet{corvi2023detection}, \citet{ojha2023towards}, and \citet{wang2020cnn}, while we retain the same weights for the rest of the detectors (\citet{koutlis2024leveraging}, \citet{chen2024drct} (ConvNext), and \citet{chen2024drct} (CLIP)), both for the acceptable performance and to ensure the generalization of attacks without the induced risk of a dataset bias if all detectors are trained on the D\textsuperscript{3} dataset.
After retraining, we observe improvements across all metrics for the four detectors, except for \citet{cavia2024real}, which shows limited gains. As the empirical results reported in the following show, however, high performance on unperturbed images does not imply greater robustness to adversarial attacks.

\begin{table*}[!tb]
\caption{White-box and black-box adversarial robustness evaluation (AUROC/ASR) on PGD with $\epsilon$ = 16/255 perturbation. Adversarial attacks are ran against the model in each row and evaluated on models in each column. † refers to detectors trained on the D\textsuperscript{3} training set. Final row $N-$  \textit{model} refers to ensemble attacks targeting all models, excluding the evaluated against model. \textbf{Bold} values corresponding to white-box evaluation in single detector attacks.}
\label{tab:whitebox_eval_16}
\centering
\vspace{0.1cm}
\setlength{\tabcolsep}{.5em}
\small
\begin{tabular}{lccccccc}
\toprule
 & Ca24\textsuperscript{†} & Ch24C & Ch24CN & Co23\textsuperscript{†} & K24 & O23\textsuperscript{†} & W20\textsuperscript{†}\\
\midrule
    Ca24\textsuperscript{†} & \colorbox{GrayLight}{\textbf{0.00/100.0}} & 0.70/0.6 & 0.85/5.9 & 0.90/7.6 & 0.68/18.5 & 0.87/4.3 & 0.94/4.2 \\
    Ch24C & 0.58/27.8 & \colorbox{GrayLight}{\textbf{0.00/100.0}} & 0.49/46.3 & 0.93/3.6 & 0.04/95.4 & 0.08/91.6 & 0.94/4.3 \\
    Ch24CN & 0.52/34.5 & 0.59/17.0 & \colorbox{GrayLight}{\textbf{0.03/96.9}} & 0.89/7.7 & 0.61/26.9 & 0.88/3.4 & 0.95/3.6 \\
    Co23\textsuperscript{†} & 0.32/59.4 & 0.65/7.5 & 0.08/90.8 & \colorbox{GrayLight}{\textbf{0.00/100.0}} & 0.49/41.4 & 0.82/9.9 & 0.01/99.3 \\
    K24 & 0.54/31.8 & 0.08/89.3 & 0.43/52.4 & 0.93/4.4 & \colorbox{GrayLight}{\textbf{0.05/93.7}} & 0.20/77.6 & 0.94/3.8 \\
    O23\textsuperscript{†} & 0.57/28.6 & 0.03/95.3 & 0.71/21.6 & 0.93/4.2 & 0.06/92.4 & \colorbox{GrayLight}{\textbf{0.00/100.0}} & 0.93/5.5 \\
    W20\textsuperscript{†} & 0.35/56.5 & 0.69/2.2 & 0.31/65.9 & 0.06/93.3 & 0.54/35.2 & 0.84/7.3 & \colorbox{GrayLight}{\textbf{0.00/100.0}} \\
\midrule
   \rowcolor{ourcolor}
    $N-$ \textit{model} & 0.36/54.7 & 0.02/97.6 & 0.20/78.2 & 0.13/87.0 & 0.05/94.3 & 0.12/86.5 & 0.26/73.5 \\
\bottomrule
\end{tabular}
\end{table*}
\begin{table*}[!tb]
\caption{White-box and black-box adversarial robustness evaluation (AUROC/ASR) on PGD with $\epsilon$ = 32/255 perturbation. Adversarial attacks are ran against the model in each row and evaluated on models in each column. † refers to detectors trained on the D\textsuperscript{3} training set. Final row $N-$  \textit{model} refers to ensemble attacks targeting all models, excluding the evaluated against model. \textbf{Bold} values corresponding to white-box evaluation in single detector attacks.}
\label{tab:whitebox_eval_32}
\centering
\vspace{0.1cm}
\setlength{\tabcolsep}{.5em}
\small
\begin{tabular}{lccccccc}
\toprule
 & Ca24\textsuperscript{†} & Ch24C & Ch24CN & Co23\textsuperscript{†} & K24 & O23\textsuperscript{†} & W20\textsuperscript{†}\\
\midrule
    Ca24\textsuperscript{†} & \colorbox{GrayLight}{\textbf{0.00/100.0}} & 0.74/$-$4.5 & 0.76/15.4 & 0.88/9.3 & 0.72/13.4 & 0.87/4.0 & 0.92/6.0 \\
    Ch24C & 0.54/32.3 & \colorbox{GrayLight}{\textbf{0.00/100.0}} & 0.35/61.1 & 0.92/4.8 & 0.01/98.7 & 0.03/97.1 & 0.92/6.3 \\
    Ch24CN & 0.48/39.7 & 0.55/21.9 & \colorbox{GrayLight}{\textbf{0.01/99.2}} & 0.87/9.7 & 0.57/31.8 & 0.86/4.9 & 0.94/4.3 \\
    Co23\textsuperscript{†} & 0.26/67.2 & 0.60/15.5 & 0.01/98.4 & \colorbox{GrayLight}{\textbf{0.00/100.0}} & 0.44/47.6 & 0.70/23.4 & 0.00/100.0 \\
    K24 & 0.51/35.5 & 0.08/88.2 & 0.38/57.8 & 0.92/5.2 & \colorbox{GrayLight}{\textbf{0.07/91.6}} & 0.22/75.4 & 0.93/5.2 \\
    O23\textsuperscript{†} & 0.54/32.0 & 0.01/98.7 & 0.51/43.5 & 0.91/5.9 & 0.03/96.4 & \colorbox{GrayLight}{\textbf{0.00/100.0}} & 0.90/8.5 \\
    W20\textsuperscript{†} & 0.28/64.4 & 0.67/5.8 & 0.19/79.3 & 0.01/98.5 & 0.45/46.1 & 0.77/14.7 & \colorbox{GrayLight}{\textbf{0.00/100.0}} \\
\midrule
   \rowcolor{ourcolor}
    $N-$ \textit{model} & 0.31/61.5 & 0.01/98.6 & 0.15/83.0 & 0.05/94.8 & 0.01/98.5 & 0.10/88.9 & 0.04/95.8 \\
\bottomrule
\end{tabular}
\vspace{-0.3cm}
\end{table*}

\myparagraph{White-box Attacks.} White-box attacks on individual detectors reveal a general lack of adversarial robustness. At $\epsilon=\frac{16}{255}$ (\autoref{tab:whitebox_eval_16}), the AUROC drops to 0.00 for five of the seven detectors and to 0.03 and 0.05 for \citet{chen2024drct} (ConvNeXt) and \citet{koutlis2024leveraging}, \ie, an ASR of at least 93.7\%; at $\epsilon=\frac{32}{255}$ (\autoref{tab:whitebox_eval_32}), the ASR is at least 91.6\%. Since both classes are perturbed, AUROC values close to 0 indicate a fully inverted ranking rather than chance-level behavior.

\myparagraph{Cross-model Transferability.} Adversarial images generated via white-box attacks against each detector were applied to the other six detectors, revealing that transfer from a single detector largely follows architecture families (\autoref{tab:whitebox_eval_16}). Perturbations crafted on a CLIP-based detector (Ch24C, K24, O23) transfer strongly to the other CLIP-based detectors, with the lowest ASR being 77.6\% for perturbations crafted on K24 and evaluated against O23. Conversely, they barely affect the CNN-based Co23 and W20 (ASR $\leq 5.5\%$, AUROC $\geq 0.93$). Perturbations crafted on Co23 or W20, in turn, transfer strongly to each other (ASR 99.3\% and 93.3\%) and to Ch24CN (90.8\% and 65.9\%), moderately to K24 (41.4\% and 35.2\%), and only weakly to Ch24C and O23 (ASR $\leq 9.9\%$). Perturbations crafted on Ch24CN and on the patch-based Ca24 transfer weakly to all other detectors (ASR $\leq 34.5\%$ and $\leq 18.5\%$, respectively). Consequently, every single-source attack leaves at least one target almost unaffected (lowest off-diagonal ASR $\leq 7.5\%$ in every row), so that a single surrogate is effective only if it shares the architecture family of the target.

\myparagraph{Leave-one-out Ensemble Attack.} In this setting, we find that the ensemble attack transfers strongly to every held-out detector, regardless of its architecture family, providing generalizability with respect to unseen detectors. Looking at \autoref{tab:whitebox_eval_16} (last row), the AUROC of the held-out detectors drops to between 0.02 and 0.36 (ASR 54.7--97.6\%). Even the two detectors \citet{corvi2023detection} and \citet{wang2020cnn}, which were barely affected by perturbations crafted on CLIP-based detectors, drop to an AUROC of 0.13 and 0.26, respectively (from 0.99 and 0.99). This drop in performance is further highlighted in the last row of \autoref{tab:whitebox_eval_32}, where we use a higher perturbation budget, with the AUROC dropping to between 0.01 and 0.31 (ASR 61.5--98.6\%), and to 0.05 and 0.04 for the two detectors, respectively. The ensemble does not always outperform the best single surrogate of the same family, but it does not require knowing the architecture family of the target, which is the property RAID relies on. \citet{cavia2024real} is consistently the most resistant held-out detector (ASR 54.7\% and 61.5\%).

These results motivate the creation of our RAID dataset, which comprises adversarial images generated via attacks on an ensemble of state-of-the-art detectors. Our benchmark dataset enables fast, standardized evaluation of new detectors against strong, transferable perturbations, facilitating assessment of their robustness to adversarial attacks.

\begin{table*}[!tb]
\centering
\caption{Evaluation of our attack methodology on AIGIs from additional generators. For each detector, we report AUROC and ASR on images from 11 generative models, under the $N-$ \textit{model} ensemble attack (all models excluding the evaluated one).}
\label{tab:external_generators}
\vspace{0.1cm}
\setlength{\tabcolsep}{.6em}
\small
\begin{tabular}{lcc ccc ccc ccc ccc ccc}
\toprule
Metric & Perturbation & & Ca24\textsuperscript{†} & Ch24C & Ch24CN & Co23\textsuperscript{†} & K24 & O23\textsuperscript{†} & W20\textsuperscript{†}\\
\midrule
\multirow{4}{*}{\textbf{AUROC}} 
 & Clean   & & 0.77 & 0.74 & 0.97 & 0.99 & 0.89 & 0.96 & 0.99 \\
 & 8/255   & & 0.46 & 0.06 & 0.43 & 0.32 & 0.09 & 0.24 & 0.46 \\
 & 16/255  & & 0.46 & 0.02 & 0.23 & 0.20 & 0.04 & 0.15 & 0.29 \\
 & 32/255  & & 0.40 & 0.01 & 0.19 & 0.08 & 0.02 & 0.12 & 0.06 \\
\midrule
\multirow{3}{*}{\textbf{ASR (\%)}} 
 & 8/255   & & 39.5 & 91.9 & 55.5 & 67.4 & 89.5 & 75.4 & 53.5 \\
 & 16/255  & & 39.8 & 97.1 & 75.8 & 79.9 & 95.5 & 84.6 & 70.8 \\
 & 32/255  & & 48.2 & 98.9 & 79.9 & 91.7 & 97.4 & 87.1 & 94.4 \\
\bottomrule
\end{tabular}
\vspace{-0.3cm}
\end{table*}

\myparagraph{Generalizability to AIGIs from Other Generators.} We apply the leave-one-out ensemble attack to the images of the 11 additional generators and report AUROC and ASR averaged over all of them in \autoref{tab:external_generators}.
While most detectors achieve high clean AUROC (0.74--0.99), the ASR reaches 39.8--97.1\% at $\epsilon=\frac{16}{255}$ and 48.2--98.9\% at $\frac{32}{255}$, with \citet{cavia2024real} again the most resistant. This further demonstrates that our proposed approach is effective across multiple generator families, thereby extending its applicability.

\begin{table*}[!tb]
\centering
\caption{Adversarial robustness evaluation of the four trained baseline classifiers. All detectors are trained on the D\textsuperscript{3} training set. RAID refers to the dataset of the adversarial images saved as raw tensors, while RAID Img refers to the released dataset of adversarial images saved as PNG images. ASR (\%) is the relative AUROC drop with respect to the clean D\textsuperscript{3} test set.}
\label{tab:raid_eval}
\setlength{\tabcolsep}{.3em}
\small
\begin{tabular}{lcc cc cc cc cc}
\toprule
  & & & \multicolumn{2}{c}{DINOv2} & \multicolumn{2}{c}{DINOv2-Reg} & \multicolumn{2}{c}{ViT-T} & \multicolumn{2}{c}{CoDE} \\
  \cmidrule(lr){4-5} \cmidrule(lr){6-7} \cmidrule(lr){8-9} \cmidrule(lr){10-11}
 Dataset & Perturbation & & \textbf{AUROC} & \textbf{ASR} & \textbf{AUROC} & \textbf{ASR} & \textbf{AUROC} & \textbf{ASR} & \textbf{AUROC} & \textbf{ASR} \\
\midrule
D\textsuperscript{3} & Clean & & 0.85 & -- & 0.85 & -- & 0.82 & -- & 0.89 & -- \\
\midrule
RAID & 8/255 & & 0.76 & 10.1 & 0.78 & 8.9 & 0.74 & 9.0 & 0.72 & 19.2 \\
RAID & 16/255 & & 0.73 & 14.7 & 0.75 & 12.6 & 0.70 & 14.2 & 0.65 & 27.7 \\
RAID & 32/255 & & 0.67 & 21.8 & 0.70 & 17.8 & 0.66 & 19.7 & 0.59 & 34.2 \\
\midrule
RAID Img & 8/255 & & 0.76 & 10.2 & 0.78 & 9.0 & 0.74 & 9.2 & 0.72 & 19.2 \\
RAID Img & 16/255 & & 0.73 & 14.7 & 0.74 & 12.7 & 0.70 & 14.3 & 0.65 & 27.8 \\
RAID Img & 32/255 & & 0.66 & 21.9 & 0.70 & 17.8 & 0.66 & 19.7 & 0.59 & 34.4 \\
\bottomrule
\end{tabular}
\end{table*}
\begin{table*}[!tb]
\centering
\caption{Adversarial robustness evaluation of five independently trained detectors, not part of the attack ensemble. AIDE, Effort, DDD~\citep{yang2025d3}, and OmniAID use their released checkpoints trained on GenImage (SD1.4); DDA uses its released checkpoint trained on DDA-aligned MSCOCO. RAID refers to the adversarial images saved as raw tensors, while RAID Img refers to the released dataset of adversarial images saved as PNGs. ASR (\%) is the relative AUROC drop with respect to the clean D\textsuperscript{3} test set.}
\label{tab:raid_eval_extra}
\setlength{\tabcolsep}{.3em}
\small
\begin{tabular}{lcc cc cc cc cc cc}
\toprule
  & & & \multicolumn{2}{c}{AIDE} & \multicolumn{2}{c}{Effort} & \multicolumn{2}{c}{DDD} & \multicolumn{2}{c}{OmniAID} & \multicolumn{2}{c}{DDA} \\
  \cmidrule(lr){4-5} \cmidrule(lr){6-7} \cmidrule(lr){8-9} \cmidrule(lr){10-11} \cmidrule(lr){12-13}
 Dataset & Perturbation & & \textbf{AUROC} & \textbf{ASR} & \textbf{AUROC} & \textbf{ASR} & \textbf{AUROC} & \textbf{ASR} & \textbf{AUROC} & \textbf{ASR} & \textbf{AUROC} & \textbf{ASR} \\
\midrule
D\textsuperscript{3} & Clean & & 0.78 & -- & 0.70 & -- & 0.89 & -- & 0.78 & -- & 0.93 & -- \\
\midrule
RAID & 8/255 & & 0.66 & 15.5 & 0.55 & 20.5 & 0.71 & 19.5 & 0.29 & 63.1 & 0.52 & 44.2 \\
RAID & 16/255 & & 0.60 & 22.1 & 0.56 & 20.0 & 0.67 & 24.9 & 0.15 & 81.1 & 0.34 & 62.9 \\
RAID & 32/255 & & 0.50 & 35.1 & 0.48 & 30.7 & 0.57 & 35.1 & 0.09 & 87.9 & 0.16 & 82.5 \\
\midrule
RAID Img & 8/255 & & 0.66 & 15.6 & 0.56 & 20.0 & 0.71 & 19.5 & 0.29 & 63.4   & 0.52 & 44.2 \\
RAID Img & 16/255 & & 0.60 & 22.3 & 0.56 & 20.2 & 0.66 & 25.0 & 0.15 & 81.4 & 0.34 & 63.1 \\
RAID Img & 32/255 & & 0.50 & 35.2 & 0.48 & 30.8 & 0.57 & 35.2 & 0.09 & 88.1 & 0.16 & 82.6 \\
\bottomrule
\end{tabular}
\end{table*}

\myparagraph{RAID Evaluation.}
While the previous evaluation provides a good assessment of RAID, as only one detector is missing from the ensemble, we additionally evaluate the released RAID dataset, generated with the full seven-detector ensemble, on detectors that took no part in its construction. We use the four additional baseline detectors detailed in \autoref{experimental_setup}, tested against the adversarial examples saved as tensors (RAID) and as PNG images (RAID Img). As reported in \autoref{tab:raid_eval}, RAID reduces their AUROC at every budget; at $\epsilon=\frac{16}{255}$, the ASR is 12.6--27.7\% and the AUROC remains between 0.65 and 0.75.
Saving the perturbations as PNG images does not reduce their effectiveness: across all nine detectors and budgets, the AUROC on RAID Img differs from RAID by at most 0.01 and the ASR by at most 0.5 points (Tabs.~\ref{tab:raid_eval} and~\ref{tab:raid_eval_extra}).
We further evaluate five independently trained detectors with their released checkpoints (\autoref{tab:raid_eval_extra}): AIDE~\citep{yanSanityCheckAIgenerated2025b}, Effort~\citep{effort}, DDD~\citep{yang2025d3}, OmniAID~\citep{omniaid}, and DDA~\citep{dda}. At $\epsilon=\frac{16}{255}$, AIDE, Effort, and DDD are moderately affected (ASR 20.0--24.9\%), while OmniAID and DDA are strongly affected, with their AUROC dropping from 0.78 and 0.93 to 0.15 and 0.34. Overall, even recent state-of-the-art detectors, trained independently of RAID and never seen by the attack, remain vulnerable to our black-box ensemble attack, with an ASR of 20.0--81.1\% at $\epsilon=\frac{16}{255}$, which shows that RAID provides a meaningful robustness test beyond the detectors used to construct it.
\section{Related Work}

\myparagraph{Generative Image Modeling.}
Learning to generate new samples from a high-dimensional data distribution, such as natural images, is not trivial.
Several approaches, such as autoregressive models~\citep{oordPixelRecurrentNeural2016,vandenoordConditionalImageGeneration2016}, VAEs~\citep{kingma2013auto,rezendeStochasticBackpropagationApproximate2014}, and GANs~\citep{goodfellowGenerativeAdversarialNets2014,zhuUnpairedImagetoimageTranslation2017,choiStarGANUnifiedGenerative2018,karrasProgressiveGrowingGANs2018,karrasStylebasedGeneratorArchitecture2019,karrasAnalyzingImprovingImage2020,karrasAliasFreeGenerativeAdversarial2021}, have been proposed.
While it has been shown that DMs~\citep{sohl2015deep,ho2020denoising,dhariwalDiffusionModelsBeat2021} can surpass GANs in terms of quality, the costly iterative denoising process has prevented the generation of high-resolution images.
As a remedy, LDMs~\citep{rombach2022high} perform the diffusion and denoising process in the latent space of a pre-trained VAE~\citep{kingma2013auto}.
Moreover, with the addition of cross-attention layers based on U-Net~\citep{ronneberger2015u}, the generation can be controlled by textual prompts
~\citep{nichol2022,rameshHierarchicalTextconditionalImage2022,saharia2022}.
Recent models have significantly advanced the resolution of generated images (up to 4k)~\citep{chenPIXARTWeaktostrong2025,zhangDiffusion4KUltrahighresolutionImage2025}, improved prompt following, and human preference~\citep{esserScalingRectifiedFlow2024,xieSANAEfficientHighresolution2024}.

\myparagraph{AIGI Detection Methods.}
Most early approaches for detecting AIGIs exploit visible flaws like differently colored irises~\citep{maternExploitingVisualArtifacts2019} or irregular pupil shapes~\citep{guoEyesTellAll2022}.
Since such imperfections are becoming less likely, several methods rely on imperceptible artifacts instead.
Such features include model-specific fingerprints~\citep{yuAttributingFakeImages2019} or unnatural frequency patterns~\citep{frankLeveragingFrequencyAnalysis2020,durallWatchYourUpconvolution2020,dzanicFourierSpectrumDiscrepancies2020}.
Instead of using handcrafted features, \citet{wang2020cnn} demonstrate that training a ResNet-50~\citep{heDeepResidualLearning2016} on real and generated images from ProGAN~\citep{karrasProgressiveGrowingGANs2018}, paired with strong data augmentation, suffices to detect images generated by several other GANs.
Subsequent works propose improved model architectures~\citep{gragnanielloAreGANGenerated2021} or learning paradigms~\citep{cozzolinoUniversalGANImage2021,mandelliDetectingGANgeneratedImages2022,chen2024drct}, with a particular focus on generalization~\citep{chaiWhatMakesFake2020,liuForgeryawareAdaptiveTransformer2024}.
A promising direction is the use of foundation models as feature extractors to avoid overfitting on images generated by a single class of models~\citep{ojha2023towards,cozzolinoRaisingBarAIgenerated2024,koutlis2024leveraging}.
Recently, some works have explored unique characteristics of DMs for detection, like frequency artifacts~\citep{corvi2023detection} or features obtained by inverting the diffusion process~\citep{wangDIREDiffusiongeneratedImage2023,rickerAEROBLADETrainingfreeDetection2024,cazenavetteFakeInversionLearningDetect2024}.

\myparagraph{Datasets for Evaluating the Robustness of AIGI Detectors.}
To the best of our knowledge, we are the first to propose a dataset for evaluating the \emph{adversarial} robustness of AIGI detectors.
However, several datasets exist to test their generalizability and robustness to common image degradations.
\emph{GenImage}~\citep{zhuGenImageMillionscaleBenchmark2023} is a large-scale dataset comprising 1.35~million generated images based on ImageNet~\citep{russakovskyImageNetLargeScale2015}.
\emph{WildFake}~\citep{hongWildFakeLargescaleChallenging2024} features images generated by GANs, DMs, and other generative models, sourced from platforms such as Civitai to cover a broad range of content and styles.
\cite{yanSanityCheckAIgenerated2025b} also collects images from popular image-sharing websites. However, their \emph{Chameleon} dataset consists of images misclassified by human annotators, making it particularly challenging.
Furthermore, \emph{Deepfake-Eval-2024}~\citep{chandraDeepfakeeval2024MultimodalInthewild2025} contains deepfake videos, audio, and images that circulated on social media and deepfake detection platforms in 2024.

\myparagraph{Adversarial Robustness of AIGI Detectors.}
The vulnerability of AIGI detectors to adversarial images was first explored by \citet{carliniEvadingDeepfakeimageDetectors2020}.
They demonstrate that by adding visually imperceptible perturbations to an image, the AUC of a forensic classifier can be reduced from 0.95 down to 0.0005 in the white-box and 0.22 in the black-box setting.
Subsequent work explores the applicability of attacks in practical scenarios~\citep{neekharaAdversarialThreatsDeepFake2021,hussainAdversarialDeepfakesEvaluating2021,mavali2024fake} as well as possible defenses~\citep{gandhiAdversarialPerturbationsFool2020}.
\citet{derosaExploringAdversarialRobustness2024} study the robustness of CLIP-based detectors, finding that adversarial images computed for CNN-based classifiers are not easily transferable and vice versa.
Besides the addition of adversarial noise, it has been shown that AIGI detectors can also be attacked by applying natural degradations (e.g., local brightness changes)~\citep{houEvadingDeepFakeDetectors2023} or by removing generator-specific artifacts~\citep{dongThinkTwiceDetecting2022,wesselkampMisleadingDeepfakeDetection2022}.
Other attacks leverage image generators themselves to perform semantic adversarial attacks, which adversarially manipulate a particular attribute of an image~\citep{mengAVAInconspicuousAttribute2024} that can even be controlled through a text prompt~\citep{liuInstruct2AttackLanguageguidedSemantic2023,abdullahAnalysisRecentAdvances2024}. A further line of work designs \emph{learned} anti-forensic attacks specifically against generated-image detectors, either by training an auxiliary network to remove GAN fingerprints via a contrastive objective~\citep{contrastivepert2023} or a deep-dithering generator~\citep{dithering2022}, or by exploiting spatial--frequency feature importance~\citep{dualdomain2026}. These attacks are typically tailored to a single detector and to GAN artifacts, and are therefore complementary to, but not directly usable within, the differentiable ensemble-transfer setting we adopt (see \ref{supp:other_attacks}). Most closely related to our imperceptibility analysis, \citet{backbone2025foundation} craft transferable adversarial perturbations against frozen foundation-model (CLIP/Swin/DINOv2) synthetic-image detectors and report perceptual-quality metrics; unlike that work, which attacks individual frozen backbones, RAID releases a reusable, multi-generator, multi-detector transferable benchmark.

\myparagraph{Benchmarking Transfer Attacks and Real-World Robustness.}
Orthogonal to adversarial robustness, a growing body of work benchmarks the effect of real-world dissemination and processing on AIGI detection. TrueFake~\citep{truefake2025} evaluates detectors on last-generation fakes shared through social platforms such as Facebook, X, and Telegram, while QuAD~\citep{quad2026} studies repeated recompression, resizing, cropping, and online near-duplicates. These works target \emph{natural} degradations and are complementary to RAID, which isolates the \emph{adversarial} worst case. On the attack side, \citet{brau2025transferbench} systematically benchmarks ensemble-based black-box transfer attacks across many surrogate--target configurations, and finds that a simple ensemble-averaged objective performs on par with or better than more elaborate ensemble attacks; our own comparison (\ref{supp:other_attacks}) shows that this holds only partly in our setting.
\section{Discussion} \label{discussion}
Adversarial robustness should always be evaluated when proposing new AIGI detection methods, as in current state-of-the-art detectors, it is vastly neglected in favor of an evaluation against \textit{naturally occurring} post-processing operations, such as resizing, cropping, blurring, JPEG compression, or noise. While robustness to these operations is indeed important, introducing a \textit{malicious actor} that leverages carefully crafted adversarial noise can evade detection by most methods, as highlighted in our work.
Nonetheless, the absence of a standard benchmark for comparison with detectors further contributes to the lack of evaluation against adversarial attacks in AIGI detection. 
As such, we introduce RAID to address this gap in the current literature and provide a more comprehensive solution for 
evaluating the adversarial robustness of AIGI detectors. Through the direct evaluation on RAID, first on the original AIGIs and subsequently on their adversarial counterparts with the chosen perturbation budget $\epsilon$, the authors can obtain metrics reflecting the detector's performance, in addition to its adversarial robustness. We recommend authors evaluating on RAID to report results on both the clean and adversarial splits, using $\epsilon=16/255$ and $\epsilon=32/255$.

As a control, we compare the adversarial degradation against a benign-noise baseline (Gaussian noise matched by $\ell_\infty$ norm; \autoref{sec:pgd_ablation}), which yields an ASR of at most 15.9\%, against 64.8--86.8\% for the attack, confirming that the observed drops stem from adversarial optimization rather than generic image corruption.

However, our method is not without its limitations. One key challenge is that a dataset like RAID requires frequent updates to remain relevant as new generative models emerge, and these models must be incorporated into our proposed ensemble attacks to maintain effectiveness. The dynamic nature of generative models demands a proactive approach to maintaining RAID's robustness over time, and we plan to address this by periodically extending the generator ensemble and releasing updated dataset versions.

Furthermore, using PGD in our attack algorithm requires that all ensemble detectors be end-to-end differentiable, which constrains our detector selection. Although several attacks do not have this constraint, their inclusion in the model ensemble requires a significant modification to the method. 

Additionally, our perturbations are not designed to be robust to post-processing operations, and due to the inherent restrictions on the input sizes of some detector architectures, all images are center-cropped to $256\times256$ before the attack; the perturbation covers this entire crop, which constitutes the released image. Four of the seven ensemble detectors further apply a $224\times224$ center crop as their first pre-processing step, meaning the image border is perturbed only through the gradients of the remaining three detectors, which process the full crop.

Several directions remain for future work: extending the attack to detectors with non-differentiable operations, including additional attack methods, and designing perturbations robust to post-processing transformations.
\section{Conclusions}
We introduce RAID, the first benchmark dataset of transferable adversarial images for evaluating the robustness of AIGI detectors. We employ an ensemble attack that demonstrates strong transferability against seven diverse detectors and cover images generated from four text-to-image generative models.
Our results further highlight the existing gap in current evaluations of state-of-the-art detectors, which are typically tested against natural post-processing but not against adversarial perturbations, to which they remain highly vulnerable. RAID addresses this gap by providing a simple, reproducible worst-case benchmark for evaluating adversarial robustness, enabling detection models to be screened under strong, transferable adversarial conditions.
\section*{Ethics Statement}
Research on the adversarial robustness of classifiers always poses a dual-use risk.
Our proposed attack methodology could be misused by an adversary to craft AIGIs that bypass detectors.
However, our work does not introduce any new attack methodologies that were not previously available to adversaries.
Therefore, we argue that our benchmark dataset primarily benefits defenders, enabling developers to evaluate and improve detectors' robustness.

\section*{Acknowledgments}
This work was carried out while H. Eddoubi was enrolled in the Italian National Doctorate on Artificial Intelligence run by Sapienza University of Rome in collaboration with the University of Cagliari. We acknowledge the CINECA award under the ISCRA initiative, for the availability of high-performance computing resources and support. This research has been also partially supported by the Horizon Europe projects ELSA (GA no. 101070617), Sec4AI4Sec (GA no. 101120393), and CoEvolution (GA no. 101168560); and by SERICS (PE00000014) and FAIR (PE00000013) under the MUR NRRP funded by the EU-NGEU.
\bibliographystyle{elsarticle-num-names} 
\bibliography{bibliography}

\appendix

\section{Detailed Ensemble Attack Algorithm}
In Alg.~\ref{alg:ensemble_pgd}, we report the complete procedure used to craft the adversarial images, making explicit the per-detector preprocessing, the aggregation of per-model gradients, the signed update, the projection onto the $\ell_\infty$ ball, and the clipping that keeps the image in the valid range $[0,1]^d$. The perturbation is initialized to zero ($\vct \delta \gets \vct 0$); we report a random-restart ablation in \autoref{sec:pgd_ablation}. Because the ensemble detectors require different input resolutions, each detector $m$ applies its own preprocessing $\pi_m$ (\eg, resizing and normalization); the perturbation is optimized over the entire $256\times256$ image: \citet{ojha2023towards}, \citet{koutlis2024leveraging}, and both variants of \citet{chen2024drct} center-crop it to $224\times224$, whereas \citet{cavia2024real}, \citet{corvi2023detection}, and \citet{wang2020cnn} process the full image.

\begin{algorithm}[t]
\caption{Ensemble PGD for crafting transferable adversarial examples.}
\label{alg:ensemble_pgd}
\begin{algorithmic}[1]
\Require Image $\vct x \in [0,1]^d$; detectors $\{f_{\vct \theta_m}\}_{m=1}^M$ with preprocessing $\{\pi_m\}_{m=1}^M$; budget $\epsilon$; step size $\alpha$; iterations $K$
\Ensure adversarial image $\vct x + \vct \delta$ with $\|\vct \delta\|_\infty \le \epsilon$
\State $\vct x \gets \text{center\_crop} (\vct x, 256\times256)$ \Comment{center crop for batch processing}
\State $\vct \delta \gets \vct 0$ \Comment{zero initialization}
\For{$k = 1$ \textbf{to} $K$}
    \State $\vct g \gets \dfrac{1}{M} \displaystyle\sum_{m=1}^{M} \nabla_{\vct \delta}\, \mathcal{L}_{CE}\!\left(\pi_m(\vct x + \vct \delta),\, y, \vct \theta_m\right)$ \Comment{aggregate gradients}
    \State $\vct \delta \gets \vct \delta + \alpha \,\sign(\vct g)$ \Comment{gradient ascent step}
    \State $\vct \delta \gets \mathrm{clip}_{[-\epsilon, \epsilon]}(\vct \delta)$ \Comment{project onto $\ell_\infty$ ball}
    \State $\vct \delta \gets \mathrm{clip}_{[0,1]}(\vct x + \vct \delta) - \vct x$ \Comment{keep image in valid range}
\EndFor
\State \Return $\vct x + \vct \delta$
\end{algorithmic}
\end{algorithm}

\section{Other Adversarial Attacks: Discussion and Results}
\label{supp:other_attacks}

To complement our PGD-based ensemble attack, we compare it with two widely used alternatives in the same leave-one-out setting: APGD~\citep{pmlr-v119-croce20b}, extended to the ensemble by averaging the per-model cross-entropy losses, and the Common Weakness Attack (CWA)~\citep{chen2023rethinking}, which is explicitly designed for transferability from model ensembles. All three attacks are untargeted, perturb both classes, start from zero, and use the budgets, number of iterations, and step sizes of RAID (\autoref{experimental_setup}). \autoref{tab:other_attacks} reports the ASR on each held-out detector.

APGD performs on par with PGD, with a mean ASR 1.2--1.5 points higher at every budget. CWA transfers more strongly at the smaller budgets, with a mean ASR of 85.7\% and 89.6\% at $\epsilon=8/255$ and $16/255$, against 70.6\% and 81.7\% for PGD. At $\epsilon=32/255$, the three attacks are within 1.2 points on average (88.7--89.9\%), and CWA is weaker than PGD on \citet{ojha2023towards} (70.9\% vs.\ 88.9\%).

\citet{brau2025transferbench} benchmark $13$ ensemble-based black-box transfer attacks (including momentum-, variance-reduction-, weight-averaging-, and query-refinement-based methods such as CWA, SVRE, LGV, MBA, and SASD-WS) across $17$ surrogate--target settings, and find that a simple query-free averaging of per-model losses (exactly the objective of \autoref{eq:ensemble_attack}) performs on par with or better than the more elaborate ensemble attacks. Our comparison shows that, in our setting, the ranking depends on the budget: CWA transfers more strongly at $\epsilon\leq16/255$, where it outperforms PGD on all seven held-out detectors, whereas at $\epsilon=32/255$ the three attacks perform similarly on average (88.7--89.9\%) and PGD is stronger on three detectors, most notably \citet{ojha2023towards} (88.9\% vs.\ 70.9\%). We therefore release, alongside the main PGD split on which all experiments in this paper are based, two additional splits computed with APGD and CWA.

We further note that the specialized anti-forensic attacks against AIGI detectors suggested in the literature are, by design, not well suited to constructing a transferable ensemble benchmark. The contrastive-perturbation attack~\citep{contrastivepert2023} and the deep-dithering approach~\citep{dithering2022} are \emph{learned} anti-forensic methods that train an auxiliary network (an encoder--decoder and a VQ-VAE-based simulator, respectively) against a \emph{single} target detector, and are oriented toward removing \emph{GAN} fingerprints rather than the diffusion-model traces relevant to our data; the dual-domain feature-importance attack~\citep{dualdomain2026} is likewise a single-detector spatial--frequency attack. None of these admit the differentiable, ensemble-averaged objective of \autoref{eq:ensemble_attack} that produces a fixed, transferable dataset across heterogeneous generators and detectors, which is the same differentiability constraint we discuss in \autoref{discussion}.

\begin{table*}[!tb]
\caption{Comparison with other ensemble attacks in the leave-one-out setting: ASR (\%) on each held-out detector. Attacks are computed on 1,000 images per class.}
\label{tab:other_attacks}
\centering
\small
\begin{tabular}{lcccccccccc}
\toprule
$\epsilon$ & Attack & & Ca24\textsuperscript{†} & Ch24C & Ch24CN & Co23\textsuperscript{†} & K24 & O23\textsuperscript{†} & W20\textsuperscript{†} & Mean \\
\midrule
\rowcolor{ourcolor}
\cellcolor{white} & PGD (Ours) & & 48.3 & 93.6 & 61.0 & 73.4 & 89.1 & 75.3 & 53.8 & 70.6 \\
\cellcolor{white} & APGD & & 47.6 & 94.6 & 62.5 & 74.1 & 89.6 & 75.8 & 60.0 & 72.0 \\
\multirow{-3}{*}{\textbf{8/255}} \cellcolor{white} & CWA & & 51.7 & 96.5 & 82.4 & 85.9 & 98.8 & 90.6 & 93.7 & 85.7 \\
\midrule
\rowcolor{ourcolor}
\cellcolor{white} & PGD (Ours) & & 54.7 & 97.6 & 78.2 & 87.0 & 94.3 & 86.5 & 73.5 & 81.7 \\
\cellcolor{white} & APGD & & 53.7 & 97.8 & 81.5 & 89.1 & 95.6 & 87.0 & 77.7 & 83.2 \\
\multirow{-3}{*}{\textbf{16/255}} \cellcolor{white} & CWA & & 59.9 & 98.1 & 92.5 & 94.1 & 99.6 & 87.5 & 95.0 & 89.6 \\
\midrule
\rowcolor{ourcolor}
\cellcolor{white} & PGD (Ours) & & 61.5 & 98.6 & 83.0 & 94.8 & 98.5 & 88.9 & 95.8 & 88.7 \\
\cellcolor{white} & APGD & & 64.3 & 98.8 & 86.1 & 94.4 & 98.7 & 89.7 & 97.0 & 89.9 \\
\multirow{-3}{*}{\textbf{32/255}} \cellcolor{white} & CWA & & 74.1 & 98.4 & 93.9 & 92.9 & 99.8 & 70.9 & 96.3 & 89.5 \\
\bottomrule
\end{tabular}
\vspace{-0.4cm}
\end{table*}

\section{Additional Adversarial Robustness Evaluations}
\label{supp:additional_results}

\myparagraph{Sensitivity to Image Transformations.} 
In our experiments, the attack aims to provide an approximate estimation of the adversarial robustness in worst-case settings. For this reason, we do not consider optimizing the adversarial perturbation to make it robust against any transformations. Nevertheless, this could be achieved by introducing data augmentations during attack optimization and leveraging the Expectation Over Transformation (EOT) algorithm~\citep{athalye2018synthesizing}, although we expect that optimizing against transformations would decrease the adversarial images' transferability to other never-before-seen detectors. Likely, in real scenarios, an attacker would perform a stronger, yet more costly, attack tailored to the target detector that may resist image transformations better. In this context, our dataset can quickly and costless provide valuable insights: if a model is not robust with respect to it, it will almost certainly be vulnerable to more realistic threat models~\citep{carlini2019evaluating}.

Nonetheless, to provide additional insights, we apply several post-processing operations to the adversarial images of the leave-one-out setting (1,000 images per class) before classification, and compute the ASR with respect to the clean AUROC without post-processing (\autoref{tab:diff_trans_metrics}). At $\epsilon=\frac{16}{255}$, a random horizontal flip barely affects the attack (ASR change of at most 7.9 points). JPEG compression ($q=75$) mostly preserves the attack on the CNN-based detectors (ASR 55.6--83.2\%) but weakens it on the CLIP-based ones (45.2--67.2\%, against 86.5--97.6\% without post-processing). Downscaling to $64\times64$ and random $64\times64$ crops, which also remove most of the image content, reduce the ASR to 3.2--60.9\% and 7.7--43.7\%, respectively. Larger budgets are more robust to compression (JPEG ASR 59.0--93.1\% at $\epsilon=\frac{32}{255}$).

\begin{table*}[!tb]
\centering
\caption{Leave-one-out ensemble PGD attack with a post-processing operation applied to the adversarial images before classification: ASR (\%) on each held-out detector, on 1,000 images per class. ASR is computed with respect to the clean AUROC without post-processing. Resize and random crop are followed by upsampling back to $256\times256$; random flip is horizontal with probability 0.5.}
\label{tab:diff_trans_metrics}
\setlength{\tabcolsep}{.5em}
\small
\begin{tabular}{llccccccc}
\toprule
$\epsilon$ & Post-processing & Ca24\textsuperscript{†} & Ch24C & Ch24CN & Co23\textsuperscript{†} & K24 & O23\textsuperscript{†} & W20\textsuperscript{†} \\
\midrule
\multirow{5}{*}{8/255} & No transform & 48.3 & 93.6 & 61.0 & 73.4 & 89.1 & 75.3 & 53.8 \\
 & Resize $64\times64$ & 6.2 & 9.3 & 24.5 & 56.2 & 38.0 & 11.4 & 26.5 \\
 & Random crop $64\times64$ & 9.6 & 23.9 & 26.5 & 40.7 & 44.2 & 27.7 & 31.4 \\
 & Random flip & 46.9 & 87.2 & 61.2 & 72.9 & 82.3 & 66.8 & 53.4 \\
 & JPEG ($q=75$) & 50.6 & 29.1 & 54.6 & 55.2 & 49.3 & 21.4 & 47.5 \\
\midrule
\multirow{5}{*}{16/255} & No transform & 54.7 & 97.6 & 78.2 & 87.0 & 94.3 & 86.5 & 73.5 \\
 & Resize $64\times64$ & 3.2 & 4.4 & 34.2 & 60.9 & 52.5 & 13.0 & 27.4 \\
 & Random crop $64\times64$ & 7.7 & 22.7 & 34.7 & 43.7 & 42.6 & 31.9 & 35.2 \\
 & Random flip & 54.2 & 94.1 & 78.0 & 86.4 & 91.2 & 78.6 & 72.6 \\
 & JPEG ($q=75$) & 55.6 & 55.3 & 73.9 & 83.2 & 67.2 & 45.2 & 72.8 \\
\midrule
\multirow{5}{*}{32/255} & No transform & 61.5 & 98.6 & 83.0 & 94.8 & 98.5 & 88.9 & 95.8 \\
 & Resize $64\times64$ & 4.1 & 7.5 & 45.8 & 69.3 & 59.5 & 14.9 & 39.7 \\
 & Random crop $64\times64$ & 15.3 & 25.8 & 40.7 & 43.6 & 41.1 & 40.5 & 35.3 \\
 & Random flip & 60.6 & 97.2 & 82.6 & 94.4 & 96.8 & 82.7 & 95.3 \\
 & JPEG ($q=75$) & 64.2 & 69.1 & 59.0 & 93.1 & 84.7 & 65.2 & 92.2 \\
\bottomrule
\end{tabular}%
\end{table*}

\myparagraph{Results with $\mathbf{\epsilon=8/255}$.} We evaluate the detectors under a smaller perturbation budget of $\epsilon=8/255$, commonly used in white-box evaluations but less relevant in the transfer-based scenario, where larger perturbations (\eg, $\epsilon=16/255$ or $\epsilon=32/255$) are typically required to achieve effective cross-model fooling. The results reported in \autoref{tab:whitebox_eval_8} confirm that. In the white-box setting, the AUROC drops to at most 0.03 (ASR $\geq 96.1\%$). In the leave-one-out setting, transfer is weaker than at larger budgets, with an ASR of 48.3--93.6\%, against 54.7--97.6\% and 61.5--98.6\% at $\epsilon=\frac{16}{255}$ and $\frac{32}{255}$ (\autoref{tab:whitebox_eval_16} and \autoref{tab:whitebox_eval_32}).

\begin{table*}[!tb]
\caption{White-box and black-box adversarial robustness evaluation (AUROC/ASR) on PGD with $\epsilon$ = 8/255 perturbation. Adversarial attacks are ran against the model in each row and evaluated on models in each column. † refers to detectors trained on the D\textsuperscript{3} training set. Final row $N-$  \textit{model} refers to ensemble attacks targeting all models, excluding the evaluated against model. \textbf{Bold} values corresponding to white-box evaluation in single detector attacks.}
\label{tab:whitebox_eval_8}
\centering
\vspace{0.1cm}
\setlength{\tabcolsep}{.5em}
\small
\begin{tabular}{lccccccc}
\toprule
 & Ca24\textsuperscript{†} & Ch24C & Ch24CN & Co23\textsuperscript{†} & K24 & O23\textsuperscript{†} & W20\textsuperscript{†}\\
\midrule
    Ca24\textsuperscript{†} & \colorbox{GrayLight}{\textbf{0.00/100.0}} & 0.68/3.5 & 0.89/1.7 & 0.92/4.9 & 0.65/22.0 & 0.88/2.8 & 0.96/2.4 \\
    Ch24C & 0.62/22.5 & \colorbox{GrayLight}{\textbf{0.00/100.0}} & 0.66/27.5 & 0.94/2.9 & 0.04/95.2 & 0.09/90.5 & 0.96/2.4 \\
    Ch24CN & 0.57/28.0 & 0.60/15.0 & \colorbox{GrayLight}{\textbf{0.01/99.2}} & 0.91/6.3 & 0.71/14.2 & 0.88/2.9 & 0.95/2.7 \\
    Co23\textsuperscript{†} & 0.36/55.1 & 0.72/$-$2.2 & 0.12/86.7 & \colorbox{GrayLight}{\textbf{0.00/100.0}} & 0.56/33.2 & 0.86/4.9 & 0.01/99.1 \\
    K24 & 0.58/27.8 & 0.10/85.2 & 0.61/32.9 & 0.93/4.4 & \colorbox{GrayLight}{\textbf{0.03/96.1}} & 0.24/73.3 & 0.95/3.3 \\
    O23\textsuperscript{†} & 0.60/24.3 & 0.03/96.2 & 0.85/6.3 & 0.94/3.2 & 0.05/93.7 & \colorbox{GrayLight}{\textbf{0.00/100.0}} & 0.95/3.1 \\
    W20\textsuperscript{†} & 0.41/48.8 & 0.72/$-$1.9 & 0.39/56.6 & 0.09/90.7 & 0.68/18.0 & 0.88/3.0 & \colorbox{GrayLight}{\textbf{0.00/100.0}} \\
\midrule
   \rowcolor{ourcolor}
    $N-$ \textit{model} & 0.41/48.3 & 0.05/93.6 & 0.35/61.0 & 0.26/73.4 & 0.09/89.1 & 0.22/75.3 & 0.45/53.8 \\
\bottomrule
\end{tabular}
\vspace{-0.3cm}
\end{table*}

\begin{table*}[!tb]
\centering
\caption{Leave-one-out ensemble PGD attack on each held-out detector at $\epsilon=16/255$ ($K=10$, $\alpha=0.03$): mean $\pm$ standard deviation over three seeded random starts, on 1,000 test images. ASR (\%) is the relative AUROC drop with respect to the clean images.}
\label{tab:loo_seed_variability}
\setlength{\tabcolsep}{.4em}
\small
\begin{tabular}{lccccccc}
\toprule
Metric & Ca24\textsuperscript{†} & Ch24C & Ch24CN & Co23\textsuperscript{†} & K24 & O23\textsuperscript{†} & W20\textsuperscript{†} \\
\midrule
AUROC & 0.35\scriptsize{$\pm$0.00} & 0.02\scriptsize{$\pm$0.00} & 0.24\scriptsize{$\pm$0.00} & 0.16\scriptsize{$\pm$0.00} & 0.06\scriptsize{$\pm$0.01} & 0.16\scriptsize{$\pm$0.01} & 0.33\scriptsize{$\pm$0.01} \\
ASR & 56.1\scriptsize{$\pm$0.6} & 96.8\scriptsize{$\pm$0.1} & 72.9\scriptsize{$\pm$0.4} & 83.8\scriptsize{$\pm$0.2} & 92.7\scriptsize{$\pm$1.0} & 82.5\scriptsize{$\pm$1.4} & 66.2\scriptsize{$\pm$0.5} \\
\bottomrule
\end{tabular}%
\end{table*}

\myparagraph{Variance Across Random Starts.}
We assess the variability of the leave-one-out ensemble attack with respect to its initialization by repeating it with three seeded random starts at $\epsilon=\frac{16}{255}$ with the selected configuration in~\autoref{sec:pgd_ablation} ($K=10$, $\alpha=0.03$), on the 1,000 images of the ablation. \autoref{tab:loo_seed_variability} reports the mean and standard deviation of the AUROC and ASR on each held-out detector: the standard deviation is at most 0.01 in AUROC and 1.4 points in ASR, so the results do not depend on the random start.

\section{Evaluation on Commercial Detectors}
\label{app:commercial}

\begin{table*}[!tb]
\caption{Performance on commercial AIGI detectors. We report mean detector scores on AIGIs, with adversarial variants evaluated on RAID examples. Metrics on real images are also included.}
\label{tab:commercial_eval}
\centering
\vspace{0.1cm}
\setlength{\tabcolsep}{.5em}
\small
\begin{tabular}{lcccccccc}
    \toprule
    Detector & & \textbf{Score} & \textbf{Score (Real)} & \textbf{Adversarial Score} & \textbf{Acc} & \textbf{Acc (Real)} & \textbf{Adversarial Acc}\\
    \midrule
    Sightengine & & 0.65 & 0.99 & 0.01 & 0.66 & 1.00 & 0.00 \\
    HIVE & & 0.59 & 0.99 & 0.45 & 0.60 & 1.00 & 0.44 \\
    \bottomrule
\end{tabular}
\vspace{-0.2cm}
\end{table*}

\begin{figure*}[t]
    \centering
    \includegraphics[width=\textwidth]{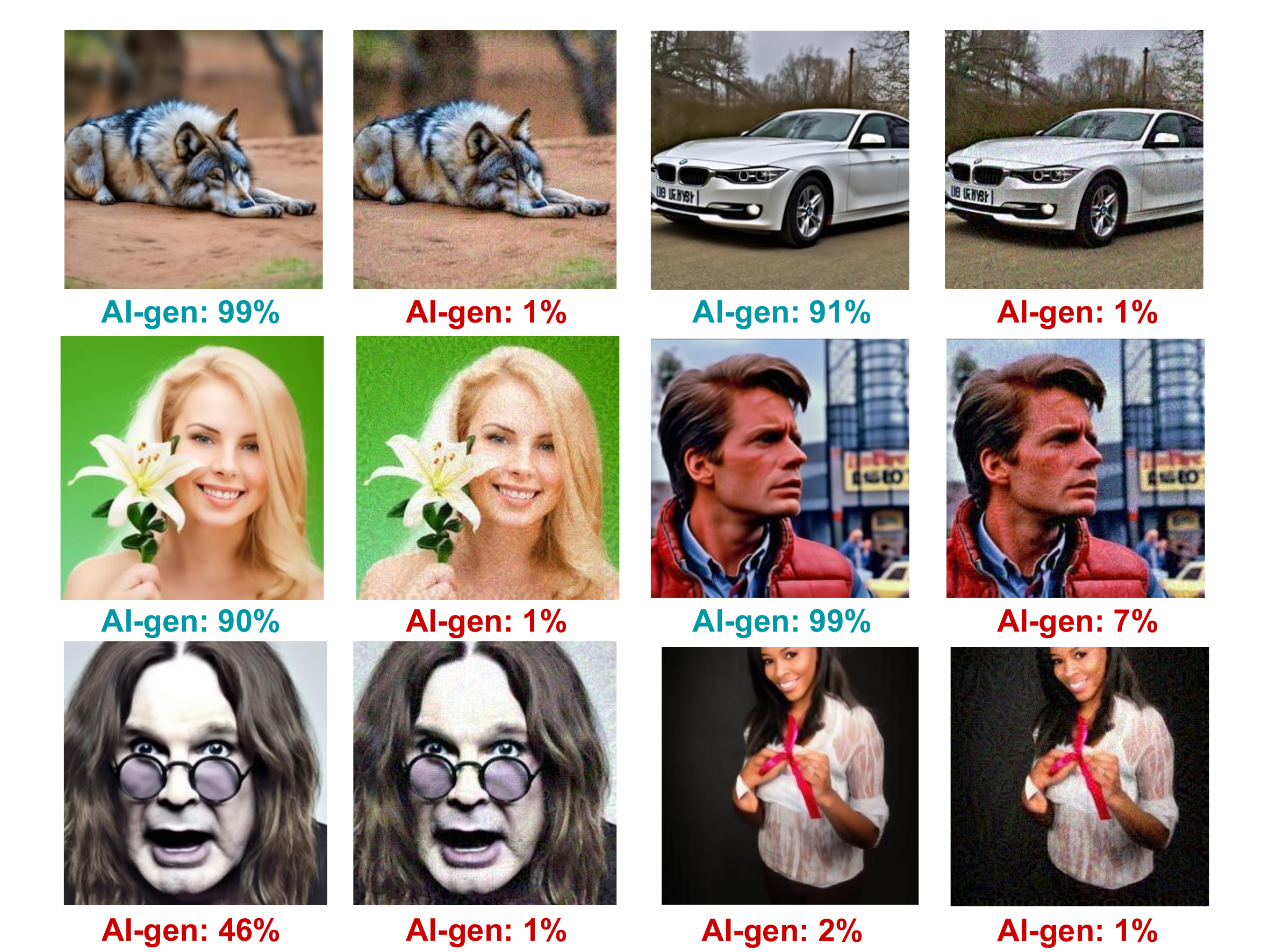}
    \vspace{-0.4em}
    \caption{Detection scores returned by Sightengine detector on a subset of clean and adversarial images. Higher scores indicate higher confidence that the image is AI-generated.}
    \label{fig:sightengine}
\end{figure*}

\begin{figure*}[t]
    \centering
    \includegraphics[width=\linewidth]{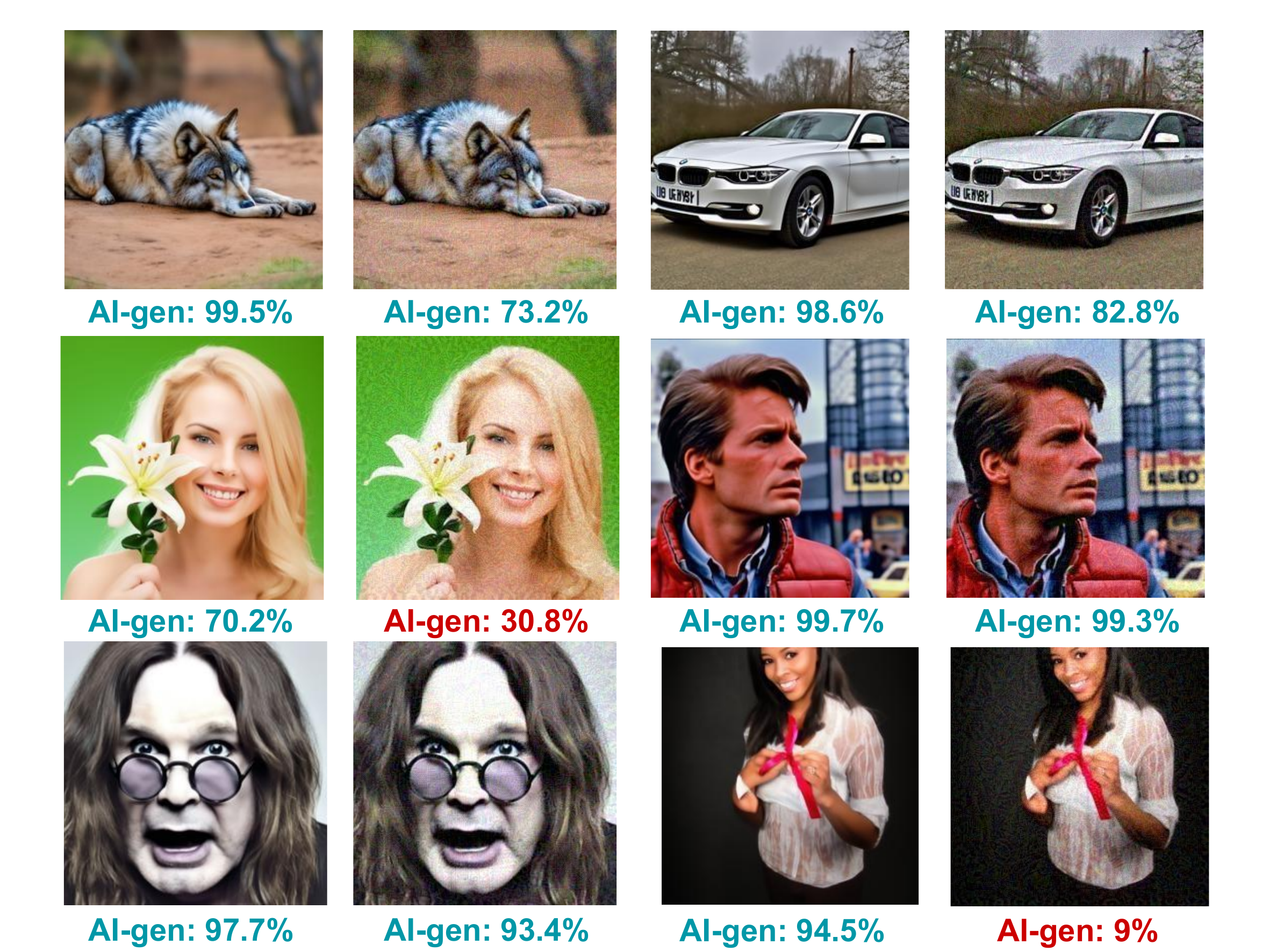}
    \vspace{-0.4em}
    \caption{Detection scores returned by HIVE Moderation detector on a subset of clean and adversarial images. Higher scores indicate higher confidence that the image is AI-generated.}
    \label{fig:hive}
\end{figure*}

We use two commercial AIGI detectors provided by Sightengine\footnote{\url{https://dashboard.sightengine.com/ai-image-detection}} and HIVE\footnote{\url{https://hivemoderation.com/ai-generated-content-detection}} to evaluate RAID in real-world settings. We report the results of a set of 50 real images and 50 pairs of clean and adversarial images in \autoref{tab:commercial_eval}. While these results are preliminary due to the dataset size, they still suggest that RAID remains effective in assessing the adversarial robustness of detectors deployed in commercial detection APIs. We compute Accuracy using a 0.5 threshold on the detectors' confidence scores. We also provide a few examples of the detection scores returned by commercial detectors on randomly selected images in \autoref{fig:sightengine} and \autoref{fig:hive}.

\end{document}